# Recognition and Co-Analysis of Pedestrian Activities in Different Parts of Road using Traffic Camera Video

Weijia Xu [1], Heidi Ross[2], Joel Meyer [3], Kelly Pierce[1], Natalia Ruiz Juri[2], Jennifer Duthie[3]

[1]. Texas Advanced Computing Center, The University of Texas at Austin, USA
[2.] Center of Transportation Research, The University of Texas at Austin, USA
[3.] Austin Transportation Department, City of Austin, USA

**Abstract**
Pedestrian safety is a priority for transportation system managers and operators, and a main focus of the Vision Zero strategy employed by the City of Austin, Texas. While there are a number of treatments and technologies to effectively improve pedestrian safety, identifying the location where these treatments are most needed remains a challenge. Current practice requires manual observation of candidate locations for limited time periods, leading to an identification process that is time consuming, lags behind traffic pattern changes over time, and lacks scalability. Mid-block locations, where safety countermeasures are often needed the most, are especially hard to identify and monitor. The goal for this research is to understand the correlation between bus stop locations and mid-block crossings, so as to assist traffic engineers in implementing Vision Zero strategies to improve pedestrian safety. In a prior work, we have developed a tool to detect pedestrian crossing events with traffic camera video using a deep neural network model to identify crossing events. In this paper, we extend the methods to identify bus stop usage with traffic camera video from off-the-shelf CCTV pan-tilt-zoom (PTZ) traffic monitoring cameras installed at nearby intersections. We correlate the video detection results for mid-block crossings near a bus stop, with pedestrian activity at the bus stops in each side of the mid-block crossing. We also implement a web portal to facilitate manual review of pedestrian activity detections by automating creation of video clips that show only crossing events, thereby vastly improving the efficiency of the human review process. The value of this work lies not with the result of this particular case study, but on a computational aided approach to conduct correlation analyses using a street view captured by a nearby traffic camera. As the methodology for the computational approach is improved, this work can be scaled to other locations or other use cases, such as determining if mid-block crossings are a result of the nearby bus stops. If a correlation is found and identified early, addressing pedestrian safety issues around mid-block crossings could prevent loss of life.

## 1. Introduction

Pedestrians are the most important and yet the most vulnerable link within any transportation system. Pedestrian safety is of critical importance for transportation system managers and operators, and safety measures require frequent revisiting due to the evolving nature of traffic patterns. Over the past years, the total number of pedestrian fatalities has been on the rise, which motivates local transportation agencies to implement additional proven countermeasures to improve pedestrian safety. Vision Zero is a holistic strategy to end traffic-related fatalities and serious injuries, while increasing safe, healthy, and equitable mobility for all. Vision Zero holds that most traffic deaths and injuries are not a result of unavoidable accidents, but are rather a preventable public health issue. This strategy was first implemented in Sweden in the 1990s and has led to major reductions of traffic deaths and serious injuries in many places around the world who have taken concrete actions to achieve its goals. As approximately 75% of pedestrian fatalities occur at mid-block locations, much focus for the Vision Zero program at the City of Austin,

Texas, is placed on strategies to facilitate safer pedestrian crossings. However, a major challenge is to proactively identify mid-block locations where those safety improvements are most needed, and to do so before incidents take place.

The collection and analysis of video data using standard CCTV pan-tilt-zoom (PTZ) traffic monitoring cameras at critical locations provides an opportunity to analyze pedestrian movements and to provide a verifiable account of road user behavior (Kastrinaki et.al., 2003; Kuciemba and Swindler, 2016). The former reduces the need to rely on ad hoc decision making (Sayed et al., 2013). However, if analyses are conducted by human observers, there is a limit to the number of locations and analysis periods that may be considered. Automated approaches to effectively recognize, analyze, and store pedestrian activities over time are needed (Zangenehpour et. al., 2015; Aubin et. al., 2015; Pusar et al., 2018). The technical challenges associated with pedestrian activity analysis using traffic monitoring video data are different from those faced when conducting traffic flow analyses. Regular roadside cameras have wide and deep fields of view. Pedestrian activities occupy only a small portion of the field, and at many locations are only present sporadically. Further, pedestrians appear smaller in size than cars and are more frequently subject to obstruction from other objects within the scene.

Incorporating Internet of Things (IoT) and smart devices within an intelligent transportation system (ITS) usually comes with substantial up-front costs for installation and deployment. At the same time, advances in algorithm development and software design bring new opportunities to increase utilization of existing transportation infrastructure. To address these challenges, we have continued developing a video processing pipeline (Xu et. al. 2018, 2019a, 2019b) to improve pedestrian crossing detections. From this pipeline, we can collect video from existing traffic monitoring cameras to automate the process of identifying pedestrian activities in the video using deep learning models. We further extend our implementation so that regions of interest within the camera view can be defined and selected to study specific types of pedestrian activities. This enables pedestrian on the side of road (e.g., at a bus stop) to be identified and inferred for activities beyond mid-block crossing events. This additional capability has the potential to help decision makers and traffic engineers gain insights on why mid-block crossings happened at a particular location.

In this paper, we present a use case on identifying pedestrian presence at bus stops and correlate that with mid-block crossing detections. The case study focused on a selected location in Austin, Texas, where mid-block bus stops located on both sides of Lamar Boulevard are visible through the City of Austin (CoA) PTZ traffic monitoring camera at a nearby intersection. We collected and studied traffic camera video footage for the first two weeks of March 2020. We isolated pedestrian activity detection to regions around each bus stop and regions in the middle of the road (to identify crossings), and correlate bus stop activity to midblock crossing events. To help traffic professionals to review and verify pedestrian activities efficiently, we have implemented a web portal with video clips that have been automatically created for each mid-block crossing identified, which users can select and review individually. Despite observed limitations, our work illustrates how the value of existing traffic monitoring camera networks can be augmented beyond everyday traffic monitoring, and used to collect valuable information on pedestrian travel patterns.

## 2. Methodological Approach and Implementation

We have developed an automated video processing framework that separates the video analysis process into two distinct parts: object recognition and analysis of the identified objects (Huang et al, 2017). We use convolutional neural networks to detect and track the motion of objects from each frame in the video stream, and then store and process information for various use cases. By combining the best practice of object recognition through deep learning and big data processing for those two parts respectively, the framework can efficiently process large-scale traffic video data automatically and meet evolving analytic needs over time.

Raw videos are originated from IP cameras in the CoA private network, which has limited accessibility. To overcome the latter, the CoA set up a proxy server to forward selected video feeds from the IP cameras to a storage cluster hosted at the Texas Advanced Computing Center (TACC). The recorded video can be then be processed by a high-performance computing cluster at TACC. Processed data is saved in a storage server, which is accessed by our project server for results dissemination purposes. The core algorithm utilizes a convolution-neural-network-based object detection system, YOLOv2, to analyze each frame of an input video (Redmon et al., 2016; Redmon and Farhadi, 2017). For each frame, the algorithm outputs a list of objects including their location in the frame, class label, and confidence of recognition. We have limited recognition to seven class labels that are most relevant, including person, car, bus, truck, bicycle, motorcycle, and traffic light. To improve algorithmic performance and maximize utilization of multi-node computing clusters, we have also adapted the YOLOv2 implementation for parallel execution (Huang et. al., 2017).

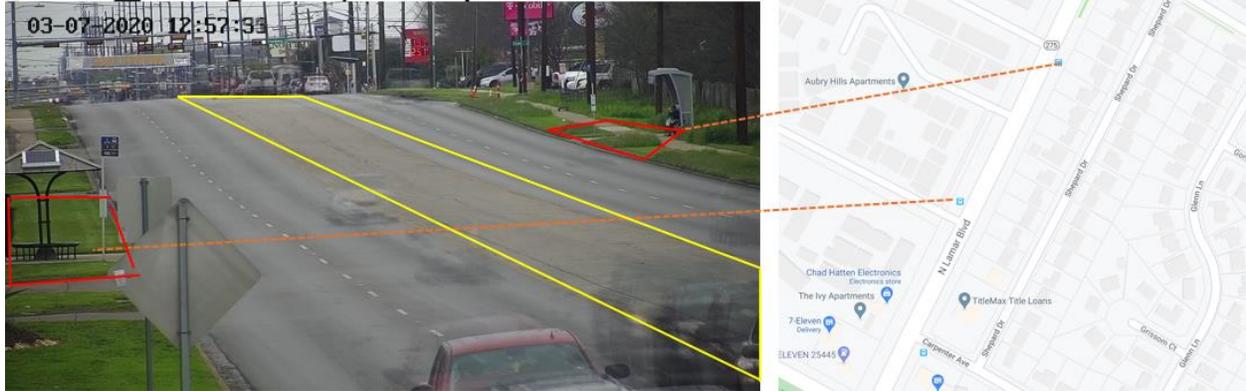

**Figure 1.** Illustration of bounding boxes and their locations on map.

To support different types pedestrian activities detection, we enable the user to define region-of-interests (ROI) over the view of the video footage. The region-of-interests can be defined as a set of arbitrary convex shapes. Multiple groups of region-of-interests can be defined and analyzed separately. Figure-1 shows an example of ROI along the center turn lane of Lamar Boulevard (yellow bounding box) and ROI on two bus stops, one on each side of the road (red bounding box).

---

Pseudo code for identify number of people waiting at bus stops.
**Input:** *D*: sequence of detections $\{f, b \mid f$: frame index, $b$: bounding box of detected object in the frame$\}$, ***ROI***: regions of interest, ***min_session_time***, ***min_no_detection***
**Output:** *S*: sequence of sessions $\{(f_b, f_e, p) \mid f_b$: begging of a session, $f_e$ end frame index of a session, p is the number of unique persons within the session$\}$

1: D → $D_{ROI}$ filter out all detections in D within region of interest specified by ROI as $D_{ROI}$
2: $D_{ROI}$ → $S_D$: merge consecutive detections in $D_{ROI}$ to form list of session which is a sequence of detections.
3: $S_D$ → $S_M$: merge sessions in $S_D$ within *min_no_detection* apart to form a new list of session $S_M$
4: $S_M$ → $S_F$: filter out sessions in $S_M$ less than *min_session_time* and store the final list of sessions in $S_F$
5: For each session in $S_F$ compute starting frame $f_b$, ending frame $f_e$, number of unique person P as ($f_b$, $f_e$, p)

---

**Figure 2.** Pseudo code to infer number of people waiting for bus from recognition results.

Since the bus stops are located on the sidewalks where other foot traffic also occurs frequently, we need to develop a process to infer the number of people waiting for a bus. We first identify time intervals (referred as *sessions*) during which people are continually detected within the regions of interests. To reduce the impact of occasional false positive identifications, a minimum time threshold, ***min_session_time*** is used so that each *session* must be longer than the specified threshold. When waiting at bus stops, people often move very little, and are easily blocked from the video camera by other objects. This can lead to false negative detection, since these people will not be detected with the described methodology. A second threshold, ***min_no_detection***, specifies a minimum time window that separates two consecutive *sessions*. If two sessions are within the ***min_no_detection*** threshold, they will be merged into one, as it is assumed the two sessions are tracking the same single person. Once sessions are defined, the number of unique persons are further inferred within each session. Figure 2 summarizes the algorithms used to infer the number of people waiting for buses.

### 3. Case Studies

The Lamar Boulevard north of Payton Gin Road location, thereafter referred as *Payton* location, was selected by the CoA Vision Zero group for a focused study on pedestrian crossing detections. There are two bus stops visible from the camera mounted at the Payton Gin Road and Lamar Boulevard intersection. The bus stops are located on Lamar Boulevard, north of the intersection. In this effort, our goal is to infer bus stop activity using video from the camera (i.e., how many people are waiting for buses throughout the day?). A view of camera and locations of two bus stops are shown in Figure 1. We have studied traffic camera videos recorded during a 14-day period from Mar 7, 2020 to Mar 19, 2020 at this location. The recording for each day is from 10AM to 8 PM. The raw recording data requires approximately 40~50 GB of storage per day, with a total of ~700GB of storage needed for the month of recorded video.

Visual summaries of person detections during the weekdays are illustrated in Figure 3. Figure 3 shows crossing detections in the Lamar Boulevard median using the video processing framework. Pedestrian crossing activities are colored by time of day, with yellow representing the AM peak period, green the midday period, blue the afternoon off-peak, and red, the PM peak period. Based on a qualitative review of the crossings shown in Figure 3, most crossings occur during the PM peak period, with midday crossings being the second busiest time of day.

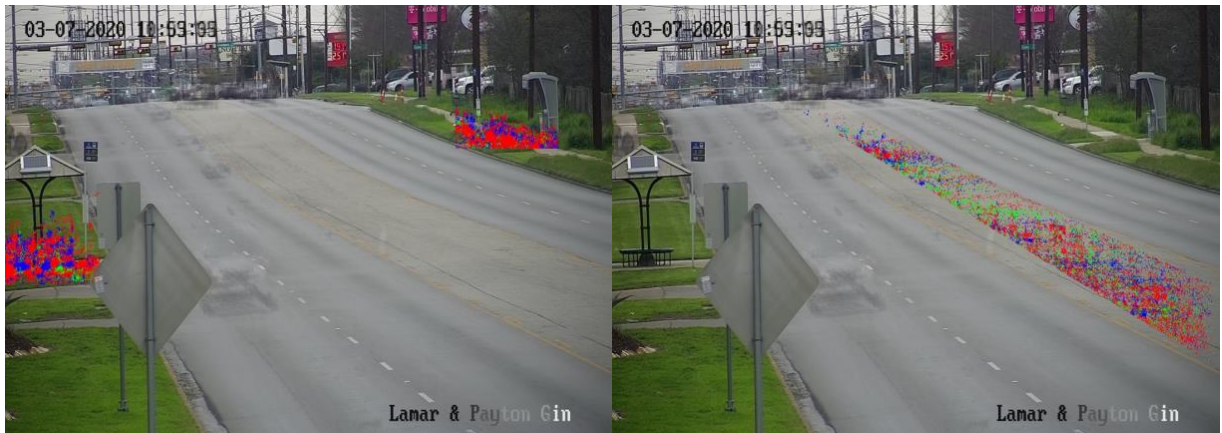

**Figure 3.** Visual summary results at Payton location during Weekdays (03/09/2020 ~ 03/13/2020 and 03/16/2020 ~ 03/19/2020). Detections at bus stops (left) and at middle of roads (right) are colored for four time periods of the day: Yellow: 7:00-10:00; Green: 10:00-13:00; Blue: 13:00-16:00; Red: 16:00-19:00.

### 3.1 mid-block crossing event inferences

We have inferred the number of mid-block crossing per hour per day, for days during which data was collected, based on the algorithm described in (Xu et. al., 2019). The result is shown in Figure 4. Figure 4 shows the inferred crossing events for each day as a stacked column. For each column, detection for each hourly block is illustrated with different colors, starting with 00:00 at the bottom and ending with 23:00 at the top. The number in each color block represents the number of inferred crossings in that particular hour. The daily crossing events inferred for this period ranges from 17 to 81 with a median of 36 crossing events per day. The results also show large variants among total crossing of different days as well as large variants in hourly inferences for some hours.

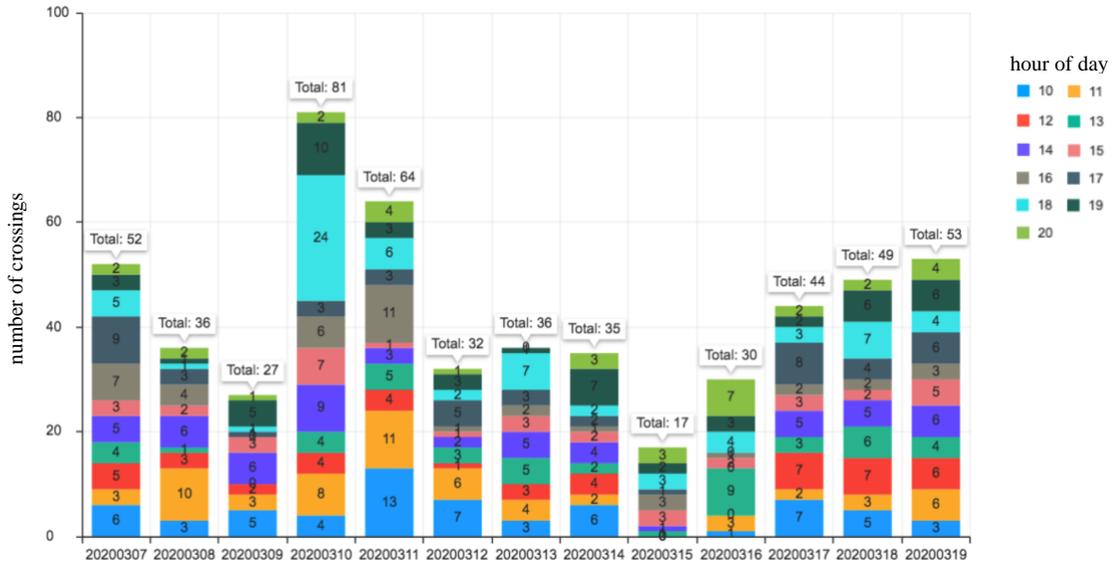

Figure 4 Number of crossing events inferred per hour per day.

Figure 5 shows a box plot of average hourly detections for pedestrian crossings by time of day at the mid-block location studied. The lines extending from the boxes indicate variability outside the upper and lower quartiles. There are 4.2 crossing events inferred on average per hour. A box plot is computed by excluding outliers to show minimum, 25% quartile, median, 75% quartile and maximum values (Figure 5). Outliers are also illustrated as circles on the plot.

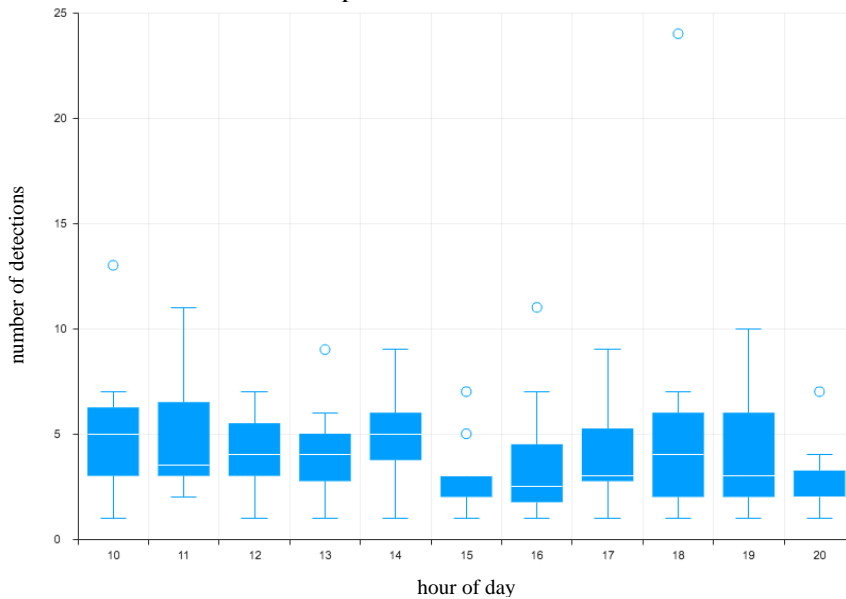

**Figure 5.** Summary of average hourly detections and outliers. Outliers are illustrated as circles on the plot. The box plot is computed by excluding outliers to show minimum, 25% quartile, median, 75% quartile and maximum values.

### 3.2 Inferences on pedestrian activity at bus stops.

The number of people waiting for buses in each hour over the recording periods are inferred according to the method described in the previous section. We choose 5 seconds as the threshold value for *min_no_detection* and 15 seconds as the threshold value for the *min_session_time*. The bus stop shown on the right side of frame (Figure 1) is the bus stop for northbound traffic (NB). The bus stop on the left side of frame (Figure 1) is the bus stop for southbound traffic (SB). Figure 7 shows the hourly number of people inferred at the NB stop for each day. The inferred total number of people at the NB stop ranges from 5 to 22 with a median of 12 for each day. However, the inferred total number of people waiting at the SB stop has a much larger variant (between 4 to 84, with a median of 33).

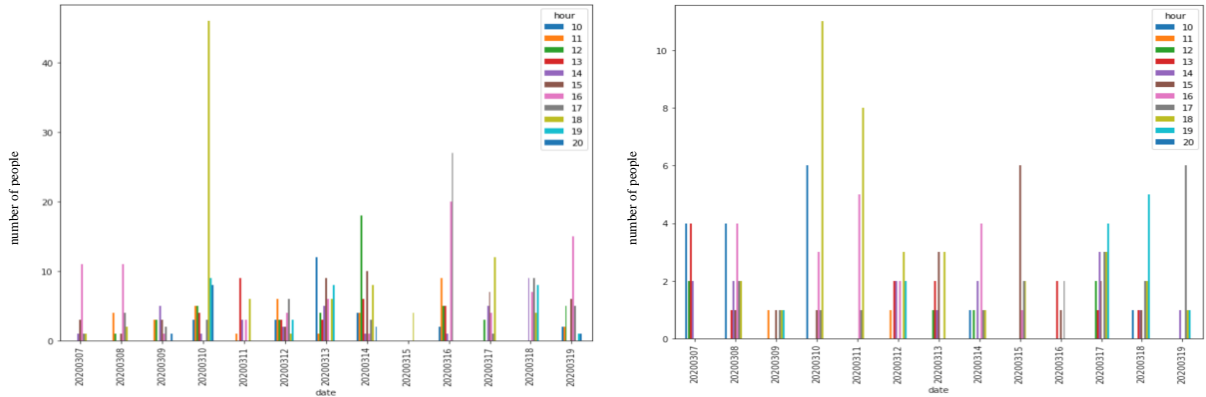

**Figure 6.** Inference of number of people waiting for bus during each hour of the day for recorded days, SB stop (left), and NB stop (right).

From Figure 6, we notice several significant spikes of the number of persons identified at bus stop area. For example, the maximum number of persons 57, was inferred for 6-7pm on Mar.10, 2020 in SB stop. Further manual review of recordings shows the inferences was due to two persons stayed at bus stop area for extended time and caused repeated detections.

In subsequent hourly pattern analysis, we incorporated an outlier detection filter to remove inferences that are clearly out of normal expectations. Further breaking down by hour, the median number of people waiting for buses per hour at the NB stop ranges from 1 to 3 across different hours of the day during this time period. For the SB stop, the median number of people waiting for buses per hour ranges from 1 to 6 across different hours of the day during this time period. When the data is combined, the median of the inferred number of people waiting at both stops per hour (across all hours and all days) is 4 (Figure 7).

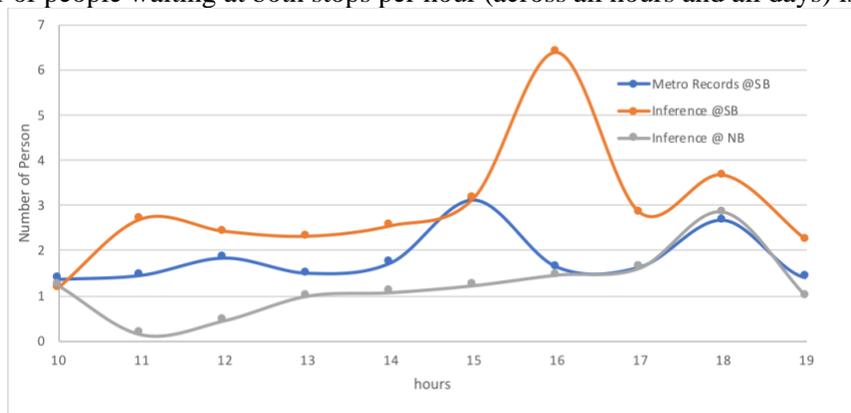

**Figure 7.** Average of inferred number of persons at bus stops for each hour during the recording periods at SB and NB stop. For comparisons, we also show the boarding records from Capital Metro at SB stop.

To check the accuracy of number of persons inferred, we also obtained boarding records at SB stop during the same time periods from the bus operator, CapMetro (Figure 7). The results show our approach inferred higher values than the boarding records from the operator's records. The mean absolute error (MAE) between our estimate with CapMetro record is 1.14. A primary reason is that our approach is based on person detected in bus stop area rather than a direct inference for number of people riding the bus which was reported Capital Metro. There are cases where person is detected around the bus stops but never boarding a bus. Manual video review also confirmed those observations.

## 3.2 Correlation between number of people waiting at bus stops and number of crossing events

To study if there are any connections between people at bus stops and inferred crossing events, a correlation analysis has been conducted for each hour of the day. In the correlation analysis, inferred values for the same hour on different days are treated as a series. Therefore, for each hour, four time series are computed for, number of crossing events (CRO), number of people waiting at the NB stop (#NB), number of people waiting at the SB stop (#SB), and the sum of people waiting at both stops (#BOTH). Each time series contains 13 data points for each day. Table 1 shows correlation results for each hour between CRO with #SB, #NB, #BOTH correspondingly. Typically, a strong correlation between two series will result a value between 0.7 ~ 1.0. Correlation values between 0.3 ~ 0.7 can be considered as moderate correlations. The results only show strong correlation (with correlation score greater than 0.7) for 18:00-19:00 at both SB and NB stop. However, this correlation may be due to the excessive false detections often observed during that hour.

**Table 1.** Correlation analysis between inferred crossing events with inferred number of people waiting at SB, NB, and both bus stops.

| Time | 10:00 | 11:00 | 12:00 | 13:00 | 14:00 | 15:00 | 16:00 | 17:00 | 18:00 | 19:00 |
|---|---|---|---|---|---|---|---|---|---|---|
| **#SB** | -0.19 | 0.20 | 0.01 | 0.39 | 0.14 | 0.05 | -0.07 | -0.36 | **0.91** | 0.40 |
| **#NB** | -0.08 | -0.03 | 0.44 | 0.37 | 0.12 | 0.10 | 0.50 | 0.14 | **0.79** | 0.06 |
| **#BOTH** | -0.22 | 0.19 | 0.07 | 0.54 | 0.18 | 0.09 | 0.09 | -0.31 | **0.92** | 0.35 |

## 3.3 A user interface for spot checking and reviewing pedestrian crossing video clips

While our approach aims to provide a robust and automated methods for detect and infer pedestrian activities, our results also indicates outliers occurs frequently among the inference results. Those outliers can be results from both model limitations and actual unexpected events on the road. To support human intervention and to gain more insights from the inference results, we have implemented a web interface for reviewing clips based on the inference results. The web interface leverages our framework developed for create web applications on cyberinfrastructure (Xu et.al., 2019).

A screenshot of this web-based interface is shown in Figure 8. At top, user may select to review results from different use cases. Summary statistics compiled using the video detection framework are shown as boxplot (Figure 5) and stack column chart (Figure 4) to give user an overview of the inference results. The boxplot on the left shows average hour pedestrian crossing events for the chosen date range, and the bar diagram on the right displays the daily number of detections for each recorded day, with each hour represented by a different color band. The graph shown through the web interfaces are interactive. Users can click on a time segment to see a list of clips associated with inferences during that time segment in the table below. The clip list table can also be interactively navigated by users to bring up clips of detections shown on the left side.

In Figure 8, the video on the bottom left shows an example of the object labeling and tracking algorithm applied to Lamar Boulevard north of Payton Gin Road in Austin, Texas. Pedestrians are identified and crossings tracked, as shown by the pink bounding boxes in this video. Note that other objects are also identified and tracked (car, bus, truck, bicycle, motorcycle, and traffic light). The data for all objects is stored with no personal identifiable information, and it can be used for future research efforts that have yet to be identified. The user can click on individual video clips to the right of the video image, to observe each pedestrian crossing. To further reducing data risks and protecting personal privacy in public, the web user interface supports authentication so that only people with account credential can access the raw video footages (Wang et.al., 2018).

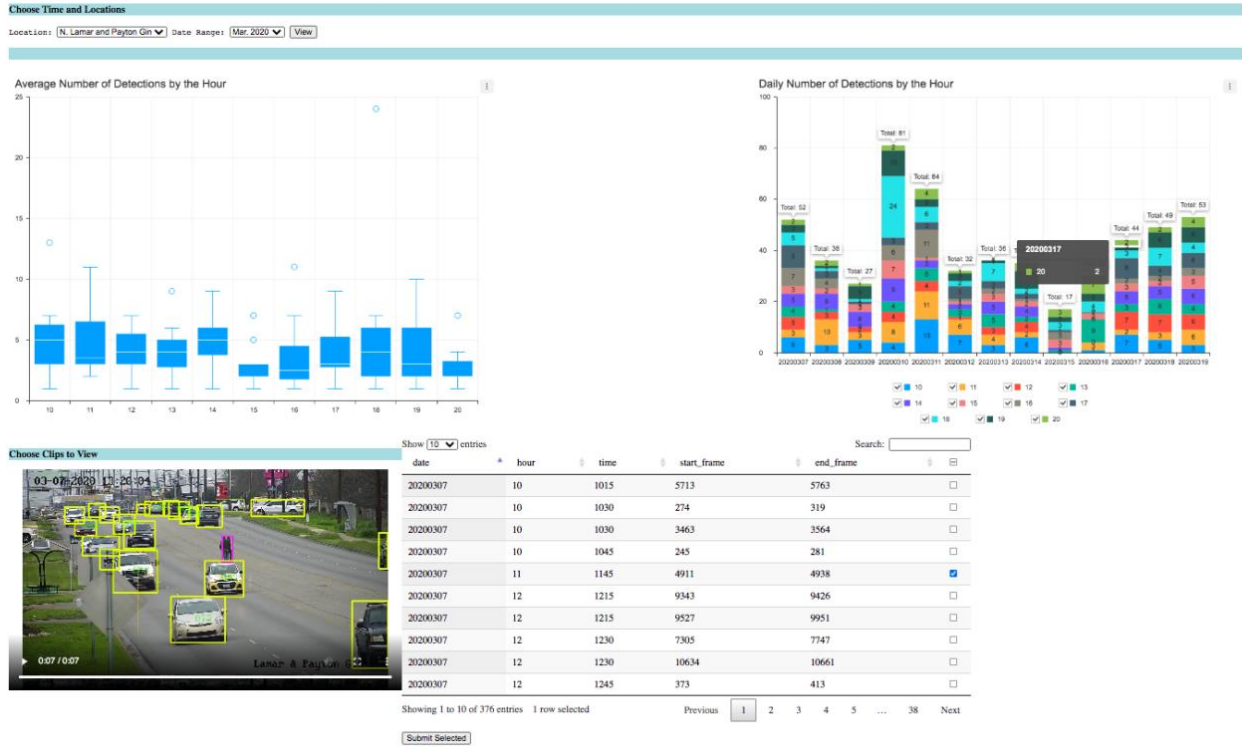

**Figure 8.** A web interface for interactive detection video reviewing purpose.

### 4. Summary and Discussion

In this paper, we presented a case study on the correlation between riders present at bus stops and mid-block crossings that occur in the area of the bus stops. Our case study leverages the latest AI technology to infer both crossing events and number of pedestrians waiting at bus stops in lieu of a manual field count. Therefore, the approach has the potential to scale to other locations and for a longer period of time than is possible using human data collection methodologies. In the case study reported here, we focused on one selected location in Austin and analyzed traffic videos collected during a period of 13 days in March 2020. We reported our inference results for pedestrian mid-block crossings and bus stop activities, and conducted a correlation analysis between these two variables for each hour of the day. Our results show no overall strong correlations are observed between person detection in bus stops with mid-block crossing for this data sets. To assist traffic engineering practitioners in reviewing and gain insights from the recording and the inference results, we also built a secure web interface to allow a user to interactively explore inference results.

The City of Austin has over 400 CCTV cameras installed at intersections in the Austin area. These cameras are commonly used for manual traffic monitoring, with no long-term recording or archiving of videos. Artificial intelligence technologies can greatly reduce the effort involved in analyzing video data, and frameworks. The framework presented here can facilitate research traditionally based on manual field and video data analysis. The intent is that this work effort will promote further work on video data applications and integration. A unique advantage of our framework is that it converts video recordings into query-able information while not saving personally identifiable information, which can accommodate multiple subsequent use cases without re-processing (Huang et al., 2017) or risks related to privacy issues.

The use cases presented in this work illustrate the benefits and limitations of the proposed methodology. Our video aggregation pipeline has the potential to support long-term pedestrian activity monitoring. The flexibility of the data selection and filtering capabilities is expected to enable further applications. In addition to the visual summaries described in this study, quantitative outputs can be generated to facilitate the comparison of conditions across different locations or time ranges, and to evaluate the impact of infrastructure changes and construction scenarios, among others. There are additional challenges in detecting pedestrian activities at bus stops. Although our motivation is to identify how many people are waiting for buses and how long they have been waiting, we also identified cases where pedestrians dwell at bus stops extended time periods that appear unrelated to bus activities. These observations can add bias to bus waiting time predictions. Furthermore, there are high numbers of duplicated detections over time, as pedestrians tend to move less while waiting for a bus to arrive.

The value of this work lies not with the result of this particular case study, but on a computational aided approach to conduct correlation analyses using a street view captured by a nearby traffic camera. As the methodology for the computational approach is improved, this work can be scaled to other locations or other use cases, such as determining if mid-block crossings are a result of the nearby bus stops. If a correlation is found and identified early, addressing pedestrian safety issues around mid-block crossings could prevent loss of life.

**Acknowledgments**

This work is based on data provided by the City of Austin, which also provided partial support for this research. The authors are grateful for this support. We would like to thank Kenneth Perrine and Chris Jordan for their help in setting up video recording environment. This work has been supported through funding from National Science Foundation (Award # 1726816). The computation of all experiments was supported by the National Science Foundation, through Frontera (OAC-1540931), and XSEDE (ACI-1953575) awards.